# AUTOMATIC DEFECT DETECTION AND CLASSIFICATION TECHNIQUE FROM IMAGE: A SPECIAL CASE USING CERAMIC TILES

G. M. Atiqur Rahaman[1]
[1]Computer Science and Engineering Discipline
Khulna University
Khulna 9208, Bangladesh
atiq99@hotmail.com

Md. Mobarak Hossain[2]
[2]Computer Science and Engineering Department
Asian University of Bangladesh
Dhaka, Bangladesh
mobarak10jan@yahoo.com

*Abstract*—*Quality control is an important issue in the ceramic tile industry. On the other hand maintaining the rate of production with respect to time is also a major issue in ceramic tile manufacturing. Again, price of ceramic tiles also depends on purity of texture, accuracy of color, shape etc. Considering this criteria, an automated defect detection and classification technique has been proposed in this report that can have ensured the better quality of tiles in manufacturing process as well as production rate. Our proposed method plays an important role in ceramic tiles industries to detect the defects and to control the quality of ceramic tiles. This automated classification method helps us to acquire knowledge about the pattern of defect within a very short period of time and also to decide about the recovery process so that the defected tiles may not be mixed with the fresh tiles.*

## I. INTRODUCTION

Image processing is one of the mostly increasing areas in computer science. As technology advances, the analog imaging is switched to the digital system now-a-days. Every day, we capture huge amount of images which are very difficult to maintain manually within a certain period of time. So the concept and application of the digital imaging grows rapidly. Digital image processing is used to extract various features from images. This is done by computers automatically without or with little human intervention. One of the most important operations on digital image is to identify and classify various kinds of defects. Thus to detect the defects from any image some methods are established and placed at three levels. At the lowest level, some techniques are available which deal directly with the raw, possibly noisy pixel values, with de-noising and edge detection being good examples. In the middle there are algorithms which utilize low level results, such as segmentation and edge linking. At the highest level are those methods which attempt to extract semantic meaning from the information provided by the lower level.

Ceramic tiles industry sector is now a very important sector for manufacturing the ceramic tiles. All production phases are technically maintained until the final stage of the manufacturing process appeared. Sometimes checking is needed for the ceramic tiles if they are able to serve customer needs, i.e. to find defected tiles. So it is an important task to categorize the ceramic tiles after production based on surface defects. The manual method of defects inspection is labor intensive, slow and subjective. Although automated sorting and packing lines have been in existence for a number of years, the complexity of inspecting tiles for damage and selecting them against the criteria of a manufacturer i.e. automated defected tiles inspection have not been possible. Again human judgment is influenced by expectations and prior knowledge. In many detection tasks for example, edge detection, there is a gradual transition from presence to absence. On the other hand, in "obvious" cases, most naive observers agree that the defect is there, even when they cannot identify the structure. Such a monitoring task is of course tedious, subjective and expensive. For all these reason no one can deny the significance of automated defect detection and classification system.

The objective of our research is to propose an efficient defect detection and classification technique which will be able to find out image defects at a high rate within a very short time.

The overall outline of this paper is mentioned as follows. Section 2 reviews the existing works briefly. Section 3 illustrates our proposed techniques. Section 4 presents the experimental results and comparison. Finally, the conclusion is presented in Section 5.

## II. EXISTING METHODS FOR DEFECT DETECTION

In the previous years, some proposed defect detection methods have been proposed to find out the image defects. But they have some limitations that can be described briefly as follows:

In [3], H. Elbehiery et al. presented some techniques to detect the defects in the ceramic tiles. They divided their method into two parts. In the first part, Existing method consisted with the captured images of tiles as input. As the output, they showed the intensity adjusted or histogram equalized image. After that, they used the output of first part as input for the second part. In the second part of their algorithm, different individual complementary image processing operations have been used in order to identify various kinds of defects. Prevailing task emphasized on the human visual inspection of the defects in the industry. But their system is not automated which is very much necessary in the manufacturing process. Again their proposed method is operation redundant because they apply their second part on every test image to identify various types of defects. Moreover, their proposed method is very time consuming.

In [4], C. Boukouvalas et al. concerned about the problem of automatic inspection of ceramic tiles using computer vision. They applied techniques for pinhole and crack detectors for plane tiles based on a set of separable line filters, through textured tile crack detector based on the wigner distribution and a novel conjoint spatial-spatial frequency representation of texture, to a color texture tile defect detection algorithm which looks for abnormalities both in chromatic and structural properties of texture tiles. But, using separate filtering techniques for different types of defects is not a good idea at all, because in such case high computational time is a major issue for applying a large number of operations. Again, their procedure is an automated visual inspection system where they only show the defects making them clear to detect the defects found on image.

In [5], Se Ho Choi et al. presented a real time defect detection method for high speed bar in coil. To enhance the performance of the detection they used edge preserving method for noise reduction, to separate images with different gray levels they used the laplacian filter ($2^{nd}$ differential) and after that they used double thresholding to binarize the image. But the major drawback of their process is that their method will not be able to find the orientation of the edge because of their using of laplacian filter, according to [7] which will be needed for the defect of ceramic tiles. The technique using the laplacian filter malfunctions for corner and curves.

We generally have found total eight types of defects from the existing defect detection methods. These types of defects are shown in the following Table I.

TABLE I. TYPES OF CERAMIC TILE DEFECTS

| Name of Defects | Description |
|---|---|
| Crack | Break down of tile |
| Pinhole | Scattered isolated black-white pinpoint spot |
| Blob | Water drop spot on tile surface |
| Spot | Discontinuity of color on surface |
| Corner | Break down of tile corner |
| Edge | Break down of edge |
| Scratch | Generally scratch on surface |
| Glaze | Blurred surface on tile |

*A. Maintaining the Integrity of the Specifications*

The template is used to format your paper and style the text. All margins, column widths, line spaces, and text fonts are prescribed; please do not alter them. You may note peculiarities. For example, the head margin in this template measures proportionately more than is customary. This measurement and others are deliberate, using specifications that anticipate your paper as one part of the entire proceedings, and not as an independent document. Please do not revise any of the current designations.

III. MATERIALS AND METHODS

*A. Proposed Approach For Defect Detection And Classification*

The complete flow chart of our proposed defect detection and classification technique has been rendered in the following Fig. 1.

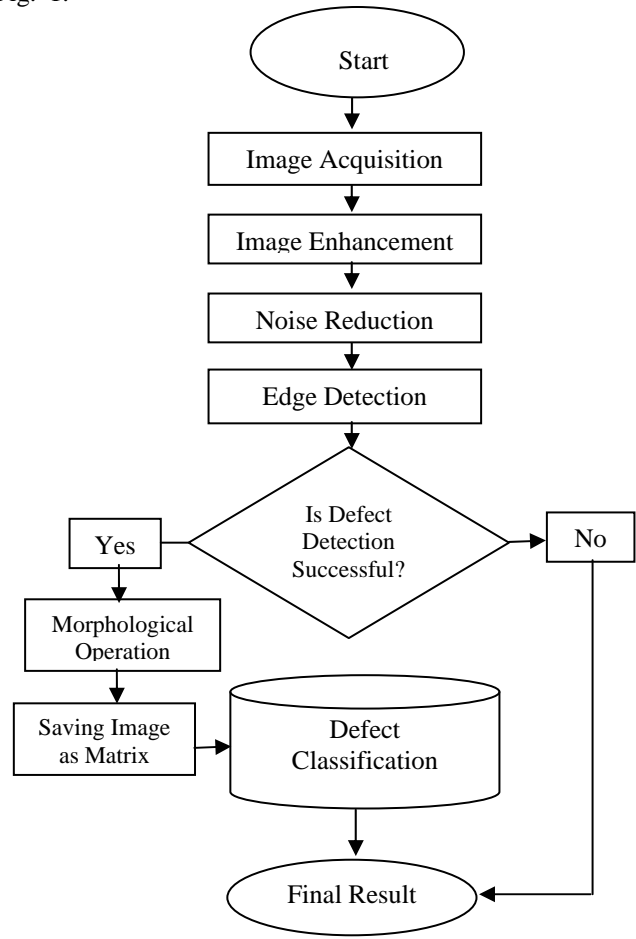

Figure 1. Flow Chart of the Proposed Method

In this section, we propose a new method for defect detection and classification. It can make the performance of defect detection rate higher than the existing one and can reduce the total computational time. Here, a common sequence of operations is applied to detect different types of defects in ceramic tiles and some proposed algorithms are employed to classify them. The proposed defect detection and classification approach will be performed using the following three steps:

**Step 1:** Performing some image preprocessing operations.

**Step 2:** Applying the proposed defect detection process.

**Step 3:** Classifying the defect using all proposed algorithms.

*B. Performing Some Image Preprocessing Operations*

At first, we have to employ some image preprocessing operations on the input image before applying the proposed defect detection process. Preprocessing steps are necessary for converting the captured RGB image found from the real source so that they can be eligible for performing any binary operations onto it. In our proposed method, we apply some preprocessing steps such as image enhancement, noise reduction etc. Again, at every preprocessing step we have a lot of options for using various kinds of methods in different applications.

*1) Image Acquisition*

Image acquisition is the process of obtaining a digitized image from a real world source. Each step in the acquisition process may introduce random changes into the values of pixels in the image which is referred to as noise. A ceramic tile image is captured and stored into the computer for further processing. This may be achieved by taking a photograph with a conventional camera, having the film made into a print and scanning the print into a computer. In our method, we have used SAMSUNG DIGIMAX-W54 digital camera for image capturing. Then, this image is trimmed with m $\times$ n (width and height) to make all the images to equal size.

*2) Image Enhancement*

Actually, image enhancement technique is to make the image clearer so that various operations can be performed easily on the image. For this, at first the captured RGB image is converted to the gray level image. Contrast stretching (often called normalization) is a simple image enhancement technique that attempts to improve the contrast in an image by stretching' the range of intensity values it contains to span a desired range of values, *e.g.* the full range of pixel values that the image type concerned allows. Low contrast images can be found due to the poor illumination, lack of dynamic range in the imaging sensor, or due to the wrong setting of the lens. The idea behind the contrast stretching is to increase the dynamic range of intensity level in the processed image.

The general process of the contrast stretching operation [1] on grayscale image is to apply the following equation on each of the pixels in the input image to form the corresponding output image pixel:

$$O(x, y) = (I(x, y) - \min)(\frac{n_i}{\max - \min}) + i$$

where, *O(x,y)* represents the output image, *I(x,y)* represents the $x_{th}$ pixel in the $y_{th}$ column in the input image. In this equation, $n_i$ represents the number of intensity levels, *i* represents the initial intensity level, *"min"* and *"max"* represent the minimum intensity value and the maximum intensity value in the current image respectively. Here *"no. of intensity levels"* shows the total number of intensity values that can be assigned to a pixel. For example, normally in the gray-level images, the lowest possible intensity is 0, and the highest intensity value is 255. Thus *"no. of intensity levels"* is equal to 256. The contrast stretching operation is applied on the grayscale images in two passes. In the first pass the algorithm calculates the minimum and the maximum intensity values in the image, and in the second pass through the image, the above formula is applied on the pixels.

In the proposed method, we enhance the gray level image to improve its visual quality and machine recognition accuracy using the following formula, described in [1]:

$$G = INTRANS(F', stretch', M, E)$$

Here, *INTRANS* performs the intensity or gray level transformations and *G* computes a contrast stretching transformation using the following MATLAB expression:

$$Contrast = 1./(1 + (M./(F + eps)).\wedge E)$$

where, parameter *M* must be in range [0,1]. The default value for *M* is $mean2(im2double(F))$ and the default value for *E* is 4. Here, *F* is gray-level image and *M* is such result which is found by applying image double and median filtering operation on *F*. *eps* returns the distance from 1.0 to the next largest double-precision number, i.e. $eps = 2\wedge(-52)$.

*3) Noise Reduction*

Noise reduction is a process of removing noise from a captured image. To remove noise some filtering techniques [1] can be proposed as follows:

One method to remove noise is by convolving the original image with a mask that represent a low-pass filter or smoothing operation. For example, the Gaussian mask comprises elements determined by a Gaussian function. This convolution brings the value of each pixel into closer harmony with the values of its neighbors. In general, a smoothing filter sets each pixel to the average value, or a weighted average, of itself and its nearby neighbors; the Gaussian filter is just one possible set of weights. But smoothing filters tend to blur an image, because pixel intensity values that are significantly higher or lower than the surrounding neighborhood would "smear" across the area. Because of this blurring, linear filters are seldom used in practice for noise reduction.

For the above reason, we proposed to use a non-linear filter which is called median filter. It is very good at preserving image detail if it is designed properly. To run a median filter:

a) Consider each pixel in the image

b) Sort neighboring pixels into order based upon their intensities

c) Replace the original value of the pixel with the median value from the list

A median filter is a rank-selection (RS) filter, a particularly harsh member of the family of rank-conditioned rank-selection (RCRS) filters [2]; a much milder member of that family, for example one that selects the closest of the neighboring values when a pixel's value is extremely in its neighborhood, and leaves it unchanged otherwise, is sometimes preferred, especially in photographic applications.

Median filter technique is good at removing salt and pepper noise from an image, and also causes relatively little blurring of edges, and hence is often used in computer vision applications.

*4) Edge Detection*

An edge may be regarded as a boundary between two dissimilar regions in an image. These may be different surfaces of the object, or perhaps a boundary between light and shadow falling on a single surface. In principle, an edge is easy to find since differences in pixel values between regions are relatively easy to calculate by considering gradients. Many edge extraction techniques [3] can be broken up into two distinct phases:

- Finding pixels in the image where edges are likely to occur by looking for discontinuities in gradients.

- Linking these edge points in some way to produce descriptions of edges in terms of lines, curves etc.

For the proposed method, we detect edge using sobel edge detection method [6] upon the resulting image. Actually there are many kinds of edge detectors. We use first derivative edge detector (sobel) to detect edges of the image. Because, it's calculation is very simple and fast to detect edges. On the other hand, if we use second derivative edge detector operator such as laplacian of gaussian operator then we will not be able to find the orientation of the edge because of using the laplacian filter. Again, if we use other kinds of gaussian edge detectors such as canny, shen castan, boie-cox operators then the operation is more complex [7].

*C. Applying the Proposed Defect Detection Process*

All preprocessing operations are applied to the reference image, stored in the computer database to compare with the test image. Let, the resulting image is $I_2$. Now we consider $I_1$ as the resulting image found from the test image after applying all preprocessing operations. We propose here a new technique. We store $I_1$ and $I_2$ as matrix form to a file. Let, these two matrices are named as $m_1$ and $m_2$. Then we count the total number of black pixels (in binary, it is represented as 1) in $m_1$ and that in $m_2$. These two are then compared. If the number of black pixels in $m_1$ is greater than the number of black pixels in $m_2$ then we can make decision that defect is found in the test image, otherwise we can say that no defect is present to the test image. To understand this concept clearly, consider the following example:

Let, we have a test image and a reference image of equal size $(5 \times 5)$. Now applying preprocessing steps on the test image we find matrix $m_1$ whose value is:

$$\begin{bmatrix} 1 & 0 & 0 & 1 & 0 \\ 0 & 0 & 1 & 0 & 0 \\ 0 & 0 & 0 & 0 & 0 \\ 0 & 0 & 1 & 0 & 1 \\ 0 & 0 & 0 & 1 & 0 \end{bmatrix}$$

Again, applying the preprocessing operations on the reference image another matrix $m_2$ is found:

$$\begin{bmatrix} 1 & 0 & 0 & 0 & 0 \\ 0 & 0 & 1 & 0 & 0 \\ 0 & 0 & 0 & 0 & 0 \\ 0 & 0 & 0 & 0 & 0 \\ 0 & 0 & 0 & 0 & 0 \end{bmatrix}$$

Here, the number of black pixels for the reference image is 2 and for the test image this number is 6. So, here obviously 6>2 and we can make decision that defect is found on the test image.

The detailed block diagram of the proposed defect detection step is shown in the following Fig. 2.

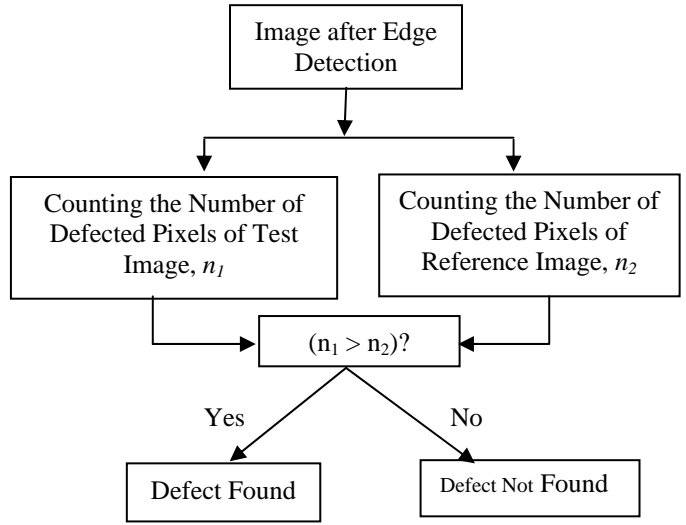

Figure 2. Block Diagram of Defect Detection Part

*D. Classifying the Defect Using All Proposed Algorithms*

In the following, our proposed detailed algorithms to classify total six kinds of defects are given step by step:

*1) Initial Steps before Classification*
*a) Plane Tiles*

**Step 1.** After finding the binary image, apply morphological operation on it.

**Step 2.** Now check every pixel elements of resulting image.

**Step 3.** If any pixel element has value 1 then

Change its value to 2.

**Step 4.** Change such coordinates of binary image as 2 which of the resulting image have value 2.

**Step 5.** Finally, save this resulting matrix into a text file.

*b) Printed Tiles*

**Step 1.** Check the reference image after processing and if any of its coordinates has value 1 then save these coordinates.

**Step 2.** After finding the binary image from test image, apply morphological operation on it.

**Step 3.** Now check every pixel elements of this resulting image.

**Step 4.** If any pixel element has value 1 then

Change its value to 2.

**Step 5.** Change such coordinates of binary test image as 2 which of the resulting image have value 2.

**Step 6.** Convert the previous saved coordinate values of binary test image as 0.

**Step 7.** Finally, save this resulting matrix into a text file.

*2) Algorithm to Determine Pinhole Defects*

Let, p_count as a variable for pinhole count, c_range as the range of corner, e_range as the range of edge and row as the maximum number of image pixels along any row and col as the maximum number of image pixels along any column.

**Step 1.** Set, temp_a =c_range , and temp_b=e_range

**Step 2.** Divide the total searching area for pinhole into three regions.

**Step 3.** For left side region,

for row consider the range from temp_a+1 to row-c_range-1

for column consider the range from temp_b+1 to c_range

(a) check every pixel coordinates whether it is 0 or not.

(b) if it is true then

(i) For each coordinate (i,j) check all of its eight neighbors.

(ii) If(i,j-1), (i,j+1), (i-1,j), (i+1,j) position values are 1 and the rest are 0, then

p_count will be incremented by 1.

**Step 4.** For right side region, range for row is from temp_a+1 to row-c_range-1 and range for column is from col-temp_a+1 to col-e_range. The rests are same as step-3.

**Step 5.** For other middle side elements, range for row is from temp_b+1 to row-e_range and range for column is from col-temp_a+1 to col-c_range. The rests are same as step-3.

**Step 6.** Finally, check value of p_count. If p_count>0, then pinhole is found, otherwise not found.

*3) Algorithm to Determine Crack Defects*

Let, c_length as the range of crack.

**Step 1.** Check every pixel coordinate (i,j) from left to right up to the last pixel element.

**Step 2.** If any (i,j) has value 1 then

(a) Consider its adjacent eight pixels and find which are 1.

(b) If any adjacent pixel has value 1 then

Current pixel coordinate will be updated to it.

(c) Apply the backtracking process to find out all connected pixels and count the length.

**Step 3.** Apply step 2 to all pixels and for each pixel find out the length of connected pixels.

**Step 4.** Counting all length of the connected pixels found from step 2 and step 3, find out the maximum number and set it to c_count.

**Step 5.** Finally, apply step 2 to specify the crack defected pixel coordinates so that other types of defects are not affected to it.

**Step 6.** If c_count > c_length, then make decision that crack is found, otherwise crack is not found.

*4) Algorithm to Determine Blob Defects*

Let, matx as size of blob, row as the maximum number of image pixels along any row and col as the maximum number of image pixels along any column.

**Step 1.** Let, start=(matx/2)+1; Here start is the middle element of (matx*matx) .

**Step 2.** Check every pixel coordinate (i,j) from left to right up to the last pixel element.

for row consider the range from start to row-start+1

for column consider the range from start to col-start+1

(a) If any pixel coordinate (i,j) is 2, then

(i) Considering it as the middle element and check the total (matx*matx) elements around it to find out how many 2 exists into these region.

(ii) Let, the total number of 2 is equal to b_length.

(iii) If b_length = (matx*matx), then

Make decision that blob defect is found and exit from loop.

(b) Otherwise, switch to next pixel coordinate at step 2.

**Step 3.** After searching every pixel coordinate, if there is no b_length matches to (matx*matx), then make decision that blob defect is not found.

*5) Algorithm to Determine Spot Defects*

Let, matx as size of spot, row as the maximum number of image pixels along any row and col as the maximum number of image pixels along any column.

**Step 1.** Let, start=(matx/2)+1; Here start is the middle element of (matx*matx) .

**Step 2.** Check every pixel coordinate (i,j) from left to right up to the last pixel element.

 for row consider the range from start to row-start+1

 for column consider the range from start to col-start+1

 (a) If any pixel coordinate (i,j) is 2, then

 (i) Considering it as the middle element and check the total (matx*matx) elements around it to find out how many 2 exists into these region.

 (ii) Let, the total number of 2 is equal to s_length.

 (iii) If s_length = (matx*matx), then

 Make decision that spot defect is found and exit from loop.

 (b) Otherwise, switch to next pixel coordinate at step 2.

**Step 3.** After searching every pixel coordinate, if there is no s_length matches to (matx*matx), then make decision that spot defect is not found.

*6) Algorithm to Determine Edge Defects*

Let, c_range as the range of corner, row as the maximum number of image pixels along any row and col as the maximum number of image pixels along any column.

**Step 1.** Initially, take a variable e_count and set its value to 0.

**Step 2.** Consider four regions in up, down, left and right side.

**Step 3.** For upper edge,

 row has fixed value 1

 for column consider the range from c_range + 1 to col-c_range + 1

 Consider each pixel coordinate (i,j).

 (i) If (i,j) coordinate has value 1 or 2, then

 Increment the value of e_count by 1 and go to step 7.

 (ii) Otherwise, continue.

**Step 4.** For lower edge, row has value row and range for column is from c_range + 1 to col-c_range + 1. The rests are same as step-3.

**Step 5.** For left edge, column has value 1 and range for row is from c_range + 1 to row-c_range + 1. The rests are same as step-3.

**Step 6.** For right edge, column has value col and range for row is from c_range + 1 to row-c_range + 1. The rests are same as step-3.

**Step 7.** If e_count > 0, then make decision that edge defect is found.

 Otherwise, edge defect is not found.

*7) Algorithm to Determine Corner Defects*

Let, row as the maximum number of image pixels along any row and col as the maximum number of image pixels along any column.

**Step 1.** Initially, take a variable c_count and set its value to 0.

**Step 2.** Check every pixel coordinates (i,j) along the range for four corner elements.

 If any coordinate has value 2, then

 Increment the value of c_count by 1.

**Step 3.** If c_count is equal to the total corner area, then

 Make decision that corner defect is found.

 Otherwise, corner defect is not found.

IV. EXPERIMENTAL RESULTS AND DISCUSSION

*A. Defect Detection*

The proposed system detects defect successfully. In this section, we show the result of step by step process of the proposed defect detection method. The defect detection rate and time efficiency are compared with the existing method [3]. We also classify here different types of defects found through defect detection process. Here it is needed to mention that during the production, many numbers of tiles are produced in industries at the same time of same colors, shape and pattern. So, all the tiles of one shape are compared with that particular type of standard tile when processing through the computers.

To understand our proposed defect detection process, we apply the proposed method on a sample RGB image for plane one colored tiles. Then we check if any kind of defect exists in this test image or not. For this, we apply our proposed preprocessing operations on this image such as image enhancement, noise reduction and edge detection. This is shown in the following Fig. 3. In this Fig. , we also show the reference RGB image for that test image and the output after applying preprocessing operations on it.

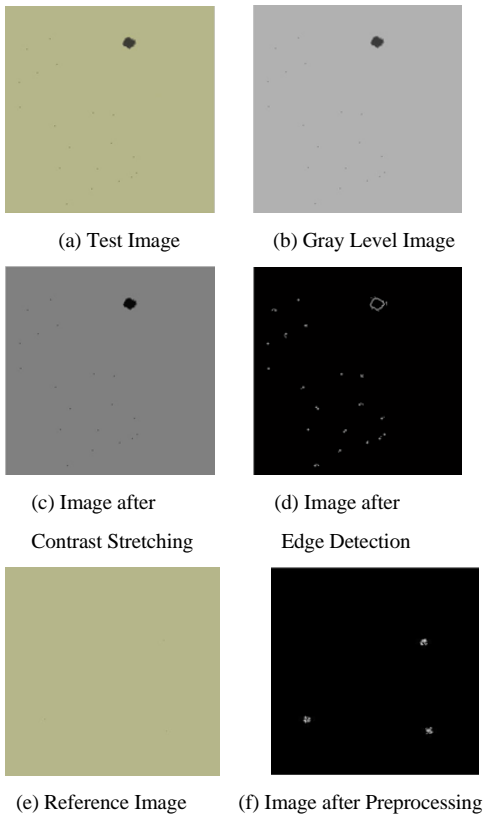

(a) Test Image  (b) Gray Level Image

(c) Image after Contrast Stretching  (d) Image after Edge Detection

(e) Reference Image  (f) Image after Preprocessing

Fig. 3: Proposed Preprocessing Steps Applied on Test Image and Reference Image

Now consider the Fig. 3(d) and Fig. 3(f). The image of Fig. 3(d) is found from the test RGB image after applying all the proposed preprocessing operations onto it. Then the total number of defected pixels is count in this image. Here the total number of defected pixels is 348. Again, the image of Fig. 3(f) is found from the reference RGB image after applying all previous operations onto it. In this image the total number of defected pixels is 105. According to our proposed defect detection method we can say, $n_1=348$ and $n_2=105$. As $n_1 > n_2$, so we can make decision that defect is found in the test image.

Again, we apply the proposed defect detection method on a sample test RGB image for printed tiles. Like before, we check if any kind of defect exists in this test image or not. Then we apply our proposed preprocessing operations on this image. This is shown in the following Fig. 4. In this Fig., we also show the reference RGB image for that test image and the output after applying preprocessing operations on it.

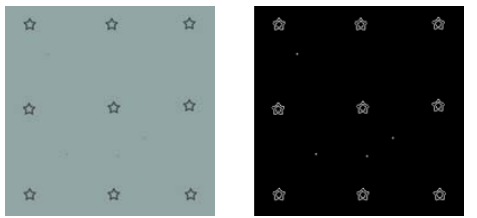

(a) Reference Image  (b) Final Image

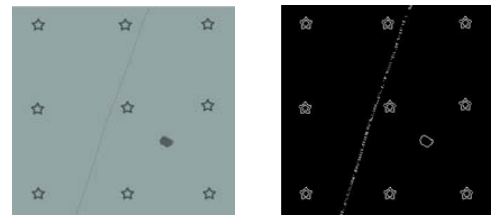

(c) Test Image  (d) Final Image

Fig. 4: (a) Reference RGB Image, (b) Final Image after Preprocessing for Reference Image, (c) Test RGB Image, (d) Final Image after Preprocessing for Test Image

Consider the Fig. 4(b) and Fig. 4(d). The image of Fig. 4(b) is found from the test RGB image after applying all the proposed preprocessing operations onto it. Then the total number of defected pixels is count in this image. Here the total number of defected pixels is 982. Again, the image of Fig. 4(d) is found from the reference RGB image after applying all previous operations onto it. In this image the total number of defected pixels is 714. According to our proposed defect detection method we can say, $n_1=982$ and $n_2=714$. As $n_1 > n_2$, so we can make decision that defect is found in the test image.

We have tested our proposed method for defect detection on about 50 ceramic tile images. In this case, the proposed defect detection efficiency is compared to the existing method [3]. We see that the detection rate for the proposed method is better than that of the existing method. Following Table II shows the efficiency comparison between the existing work and the proposed work and Fig. 5 represents the detection efficiency through a chart.

TABLE II. EFFICIENCY COMPARISON BETWEEN EXISTING WORK AND OUR WORK

| Number of Tiles | Detection Efficiency | |
|---|---|---|
| | Existing Work | Our Work |
| 10 | 90% | 100% |
| 20 | 85% | 90% |
| 30 | 86.67% | 90% |
| 40 | 87.5% | 92.5% |
| 50 | 88% | 92% |
| **Average** | **87.4%** | **93%** |

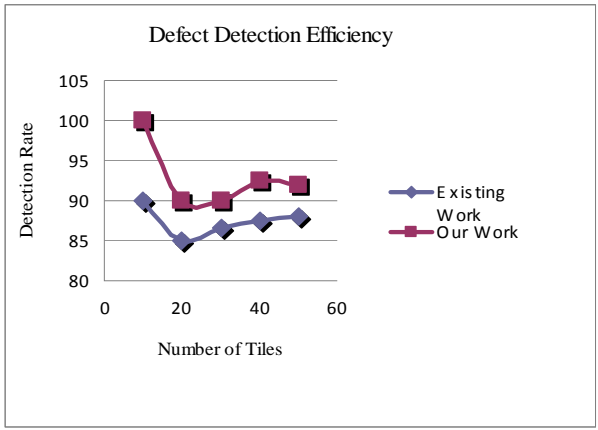

Fig. 5: Efficiency Comparison Chart between Existing Work and Our Work

We also compute the required time for the proposed method and the existing method [3]. The proposed method needs less time than the existing one. Table III shows the time comparison between the existing work and proposed work. Fig. 6 represents time comparison through a chart.

TABLE III. TIME COMPARISON BETWEEN EXISTING WORK AND OUR WORK

| Number of Tiles | Required Time (in sec.) | |
|---|---|---|
| | For Existing Method | For Our Method |
| 10 | 38.6363 | 10.3023 |
| 20 | 77.4986 | 20.4579 |
| 30 | 116.941 | 30.5645 |
| 40 | 157.204 | 40.2381 |
| 50 | 193.887 | 49.8988 |

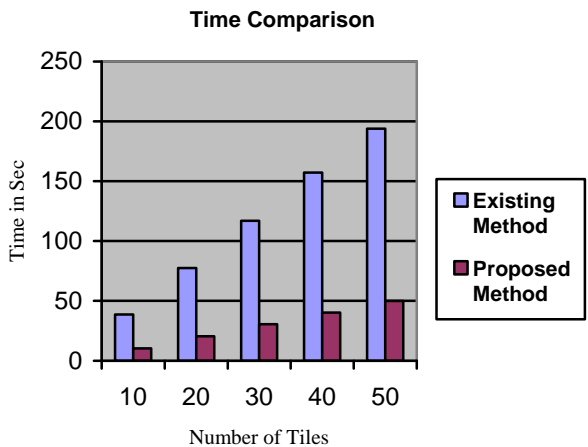

Fig. 6: Time Comparison Chart between Existing Work and Our Work

### B. Defect Classification

We need to classify the various kinds of defects after defect detection. For this purpose we store the output of the above resulting image into a file as matrix form. Then previously mentioned algorithms are applied on that matrix. In the following Table IV, we show our result after applying the proposed classification algorithms for both plane and printed tiles mentioned above.

TABLE IV. CLASSIFICATION RESULT

| Types of Defect | Result (for Plane Tiles) | Result (for Printed Tiles) |
|---|---|---|
| Pinhole | Found | Not Found |
| Crack | Not Found | Found |
| Blob | Found | Found |
| Spot | Not Found | Not Found |
| Edge | Not Found | Not Found |
| Corner | Not Found | Not Found |

Table V represents the classification efficiency of the proposed method for both plane and printed tiles and Fig. 7 represents this efficiency through a chart.

TABLE V. CLASSIFICATION EFFICIENCY OF THE PROPOSED METHOD FOR 100 TILES

| Name of Defects | Classification Rate for Both Plane and Printed Tiles (%) |
|---|---|
| Pinhole | 93.48 |
| Crack | 86.49 |
| Blob | 87.50 |
| Spot | 90.00 |
| Edge | 96.77 |
| Corner | 93.55 |

Fig. 7: Classification Efficiency Chart for Proposed Method

In this section, we have shown the result of the proposed defect detection method applied on a particular RGB image. We also have show the comparison between the existing method [3] and the proposed method in the case of detection efficiency and time efficiency. As a result, we find that our proposed method is better than the existing one. The detection rate of the proposed method is average 93% for a number of tiles, where for the existing method this rate is 87.4%. Again, the proposed method requires total 49.8988 seconds to process 50 tiles, where the existing method requires a huge time and it is total

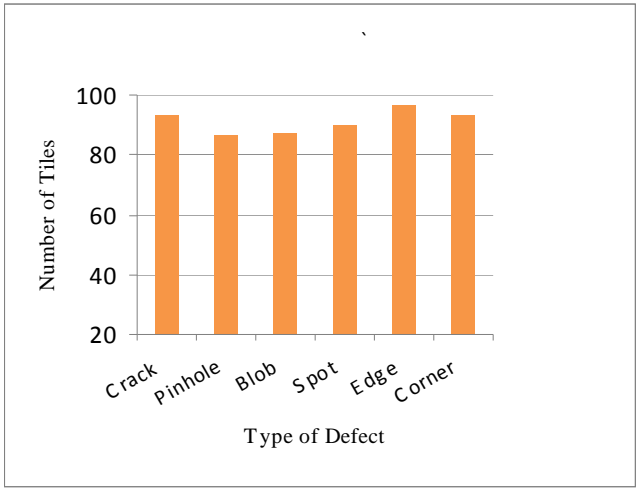

193.887 seconds. We also have shown the classification efficiency for different types of defects.

## V. CONCLUSION AND FUTURE WORK

We have proposed a defect detection method for ceramic tiles and compared our technique with the existing defect detection method. We also have shown a comparison for defect detection and processing time between the proposed method and the existing method [3]. Detection rate of the proposed method is better than that of the existing method and we need less time to detect defects than the existing one. We also have established defect classification algorithms for different types of defects. Finally, we have calculated the performance of efficiency for defect classification.

The proposed method fails to detect the glaze and scratch faults. So it may be future work to detect and classify the glaze and scratch faults. We propose no method to improve the computational time for the proposed classification technique. In this case, a future work should be done on reducing the total computational time for defect classification.

## REFERENCES

[1] R. C. Gonzalez, R. E. Woods, "Digital Image Processing", Pearson Education (Singapore), Pte. Ltd., Indian Branch, 482 F.I.E. Patparganj, 2005-2006.

[2] Puyin Liu, Hongxing Li, "Fuzzy Neural Network Theory and Application". World Scientific, 2004.

[3] H. Elbehiery, A. Hefnawy, and M. Elewa, "Surface Defects Detection for Ceramic Tiles Using Image Processing and Morphological Techniques", Proceedings of World Academy of Science, Engineering and Technology, vol 5, pp 158-160, April 2005, ISSN 1307-6884.

[4] C. Boukouvalas, J. Kittler, R. Marik, M. Mirmehdiand, M. Petrou, "Ceramic Tile Inspection for Colour and Structural Defects", under BRITE-EURAM, project no. BE5638, pp 6, University of Surrey, 2006.

[5] Se Ho Choi, Jong Pil Yun, Boyeul Seo, Young Su Park, Sang Woo Kim, "Real-Time Defects Detection Algorithm for High-Speed Steel Bar in Coil", Proceedings of World Academy of Science, Engineering and Technology, Volume 21, January 2007, ISSN 1307-6884.

[6] Se Ho Choi, Jong Pil Yun, Boyeul Seo, Young Su Park, Sang Woo Kim, "Real-Time Defects Detection Algorithm for High-Speed Steel Bar in Coil", Proceedings of World Academy of Science, Engineering and Technology, Volume 21, January 2007, ISSN 1307-6884.

[7] Mohamed Roushdi, "Comparative Study of Edge Detection Algorithms Applying on the Grayscale Noisy Image Using Morphological Filter", GVIP Journal, Volume 6, Issue 4, December, 2006.

## AUTHORS PROFILE

**G.M. Atiqur Rahaman** received B.Sc. in Computer Science and Engineering with Distinction from Computer Science and Engineering Discipline in 2003 from Khulna University, Bangladesh. Currently he is Assistant Professor of the same Discipline of the same University. He has published about 08 papers in national and international journals as well as international conference proceedings. His areas of interest include Machine Intelligence, Data Mining, Computer Vision, Digital Image Processing, Color Informatics etc.